\documentclass{article}

\usepackage{neurips_2021}

\usepackage{filecontents}
\usepackage[backref, colorlinks=true, allcolors=blue, final]{hyperref}
\usepackage[utf8]{inputenc} 
\usepackage[T1]{fontenc}    
\usepackage{url}            
\usepackage{booktabs}       
\usepackage{amsfonts}       
\usepackage{nicefrac}       
\usepackage{microtype}      
\usepackage{xcolor}         
\usepackage{algpseudocode}
\usepackage{graphicx,wrapfig,lipsum}
\usepackage{tabularx,booktabs}
\usepackage{multirow}
\usepackage[center]{caption}
\usepackage{subfigure}
\usepackage{amsmath}
\usepackage{mathtools}
\usepackage{listings}
\usepackage{color}
\usepackage{url}
\usepackage{tikz}
\usepackage{etoolbox}
\usepackage{floatrow}
\usepackage{algorithm}
\usepackage{float}

\usepackage{blindtext}

\title{Image Reconstruction using Enhanced Vision Transformer}

\begin{document}

\maketitle

\begin{abstract}
  Removing noise from images is a challenging and fundamental problem in the field of computer vision. Images captured by modern cameras are inevitably degraded by noise which limits the accuracy of any quantitative measurements on those images. In this project, we propose a novel image reconstruction framework which can be used for tasks such as image denoising, deblurring or inpainting. The model proposed in this project is based on Vision Transformer (ViT) that takes 2D images as input and outputs embeddings which can be used for reconstructing denoised images. We incorporate four additional optimization techniques in the framework to improve the model reconstruction capability, namely Locality Sensitive Attention (LSA), Shifted Patch Tokenization (SPT), Rotary Position Embeddings (RoPE) and adversarial loss function inspired from Generative Adversarial Networks (GANs). LSA, SPT and RoPE enable the transformer to learn from the dataset more efficiently, while the adversarial loss function enhances the resolution of the reconstructed images. Based on our experiments, the proposed architecture outperforms the benchmark U-Net model by more than 3.5\% structural similarity (SSIM) for the reconstruction tasks of image denoising and inpainting. The proposed enhancements further show an improvement of \textasciitilde5\% SSIM over the benchmark for both tasks.
\end{abstract}

\section{Introduction}\label{sec:intro}
Image reconstruction is an active research area where the main goal is to produce clear images in some limited environment. Digital image reconstruction involves removing noise and blur from the input images. Deblurred and denoised images are essential in applications such as healthcare where images like Magnetic Resonance Imaging (MRI) are hard to obtain and are often blurred since subjects may move during scanning, making the images hard to interpret. Recently, Vision Transformer (ViT) \cite{dosovitskiy2020image} has shown great success on image classification tasks. However, its usage for image reconstruction tasks is limited.

In this project, we propose a novel image reconstruction framework using ViT by enhancing various components of the ViT architecture. The inputs to a standard ViT are the pixel patches, called tokens, generated from the original image. These tokens are concatenated with the positional embeddings and fed to the transformer, after which attention is applied. In this project, the enhancements we propose to the ViT are as follows:
\begin{itemize}
    \item Use the SPT technique \cite{lee2021vision} to improve the tokenization process by generating overlapping tokens.
    \item Use the RoPE method \cite{jeevan2021vision} to improve the way positions are encoded with the tokens.
    \item Enhance the attention mechanism to avoid smoothing of attention scores by using a learnable temperature employing the LSA technique \cite{lee2021vision}.
    \item Use a discriminator to calculate the binary cross entropy loss of the reconstructed images to further improve the resolution of the reconstruction.
\end{itemize}
 We tested our architecture on two reconstruction tasks, image denoising and inpainting, and compared the results against the benchmark U-Net model and the baseline vanilla ViT. We also performed a comprehensive analysis of the various proposed enhancements. The proposed framework can be used for various critical applications such as MRI.

\section{Related Work}\label{sec:relatedwork}
For a computer vision system, image processing is a key component but considered as a low-level analysis of images. The results of processing visual data largely affect the high-level tasks such as image recognition or object detection. Deep learning has been widely used for a plethora of these tasks such as image super-resolution, deblurring, denoising, inpainting and colorization of images. The tasks that this project focuses on are as follows:
\begin{itemize}
    \item Denoising: Image noising is the addition of some noise function to each image pixel such as Gaussian noise, Brownian noise or impulse-valued noise \cite{lone2018noise}. These noises can be caused by sharp and sudden disturbances in image signals. Impulse-valued noise (salt and pepper noise) presents itself as sparsely occurring white and black pixels in a gray image. Denoising is the task of removing such noise.
    \item Inpainting: It is the task of reconstructing missing regions in an image \cite{guillemot2013image}. The missing/ damaged parts are filled such that the reconstructed image looks realistic. Commonly introduced paintings are vertical or horizontal bars with all pixel values replaced by 0, i.e., black pixels.
\end{itemize}
For constructing high resolution images from low-resolution data, SRCNN is a pioneering research work \cite{dong2014learning, lim2020deep}. Similarly for denoising, deep Convolutional Neural Network (CNN) architectures have been experimented with varying layer configurations and activations. Some significant works include the multi-scale CNN-based model proposed by Nah et al. \cite{nah2017deep} that uses the coarse-to-fine approach. Later, Lim et al. \cite{lim2020deep} proposed a deep spectral-spatial network that considers both spectral as well as spatial aspects in a cascade of two networks. The success of CNNs is often partially attributed to the inductive biases inherent in CNNs, allowing impressive data efficiency. U-shaped architectures are also popular for denoising tasks and serve as a benchmark for comparison. Various enhancements have since been made to the U-Net architecture such as Thesia et al. \cite{thesia2021image} used latent features of various U-Nets to determine the input image noise distribution which help with denoising.

Based on the success of transformer-based models in text processing tasks, the next important landmark in computer vision is the use of attention-based models for image processing. Some significant works include introduction of spatial attention for image augmentation \cite{yuan2021ocnet} and replacement of CNN with self-attention blocks. Wu et al. \cite{wu2020visual} proposed the use of transformer-based pre-training for image recognition and Chen et al. \cite{chen2019graph} then proposed the use of GPT model for image classification.

Transformers for vision, called vision transformers \cite{dosovitskiy2020image} are convolution-free architecture that have shown superior performance compared to the state-of-the-art CNNs for image classification when trained on millions of images. Recently, authors in \cite{tian2020attention} used attention mechanism along with convolution to account for the fact that in CNNs, as the depth of neural network increases, shallow layers start losing their effect as compared to deep layers in attention guided CNNs.

Some variants of U-Net involving efficient self-attention working on high resolution images \cite{chen2021self} have also been proposed. The authors in \cite{lee2021vision} enhanced the local attention in transformer by shifting windows, which increased the locality bias and resulted in data efficiency inspired from \cite{liu2021swin}. However, not much work has been done on using transformers for low-level image tasks like reconstruction, specifically denoising and inpainting, which are the focus of this project.

Recently, Generative Adversarial Networks (GANs) for image deblurring task are also being used often. Generative models such as DeblurGAN \cite{kupyn2018deblurgan} and conditional GANs \cite{lin2019tell} to locate and sharpen the blurred edges have shown great success. Previous studies have shown that GANs can be used to synthesize high-resolution photo-realistic images \cite{wang2018high}. Duran et al. \cite{goodfellow2014generative} used transformer-based generator and convolutional discriminator for image reconstruction. The authors showed that this approach shows better reconstructions while retaining the benefits of attention-based generator.

\section{Formal description}\label{sec:description}
This section describes the standard ViT architecture and its limitations, followed by descriptions of the various proposed enhancements for two image reconstruction tasks, image denoising and inpainting.

\begin{figure}[h]
\centering
\includegraphics[scale=0.33]{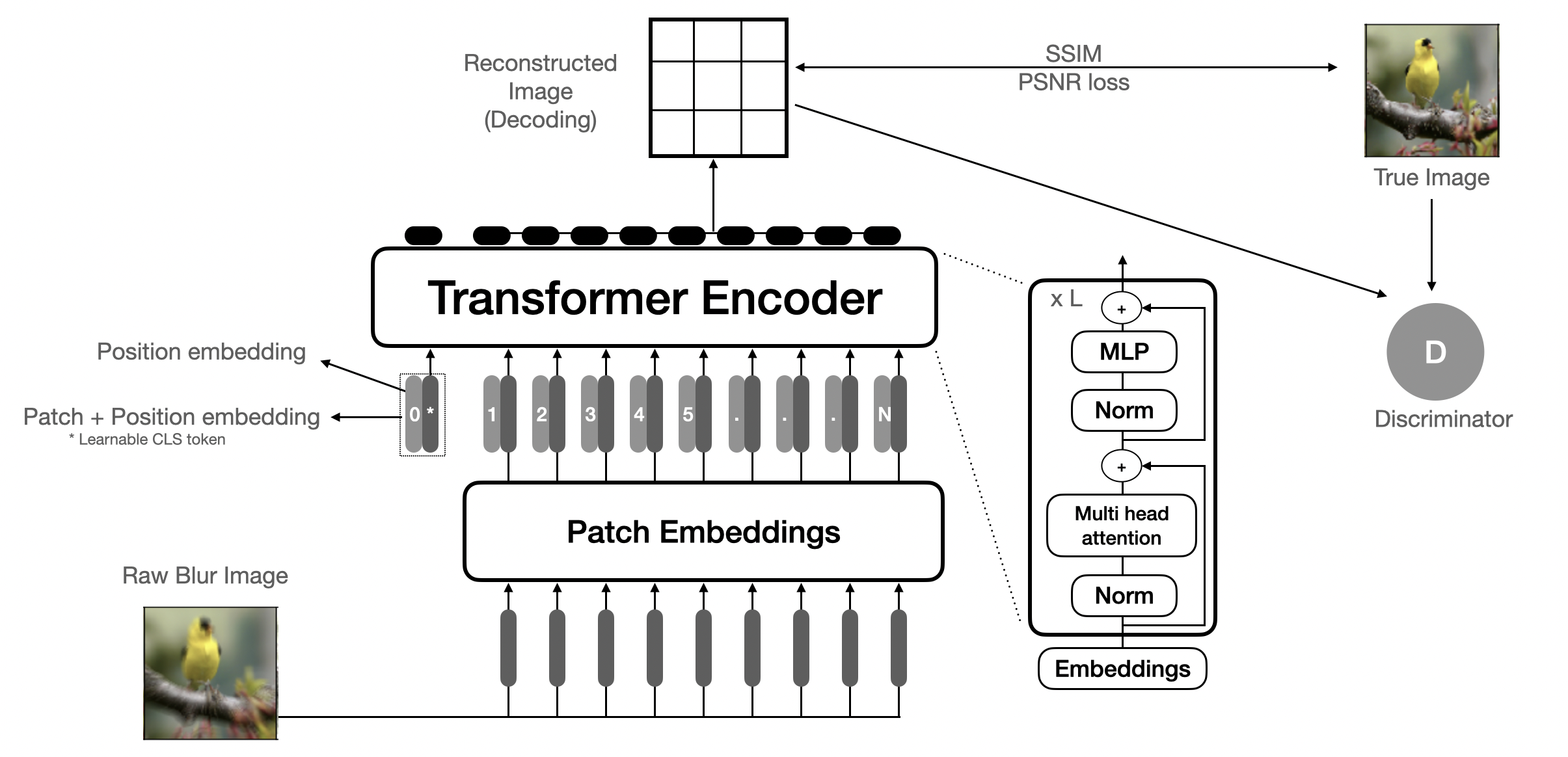}
\caption{Overview of the proposed architecture}
\label{fig:overviewViT}
\end{figure}

\subsection{Vanilla ViT}
The ViT is based on transformer architecture. For image classification, the input image is spatially divided into a sequence of N equally sized image patches. These patches are then input as a sequence of linear embeddings to the transformers, called as patch embeddings. Since the transformer encoder itself does not inherit any notion of positional information, learnable position embeddings are introduced. A learnable classification token (CLS) is prepended to the sequence of patch embeddings. These N + 1 feature vectors serve as input to the transformer encoder. For image classification task, only the output representation of the classification token is fed into a classification head, which returns the estimated class label of the input image. The final output is learned using a cross-entropy loss using softmax over the last layer’s output.

In this case, the CLS token head helps in classification and all the other context token heads generated from the actual image patches are discarded. For the task of image reconstruction, we discard the classification token as it becomes redundant for image reconstruction. The classification head is then replaced by a reconstruction head that maps the transformer output back to a visual image. The high-level architecture is shown in Fig \ref{fig:overviewViT}.

However, the problem with this ViT architecture is that it requires a lot of data and memory for effective learning. Therefore, we need to make some modifications to make use of the limited compute resources. Another problem is that ViT inherently lacks locality inductive bias. Two main reasons for this behaviour are:
\begin{itemize}
    \item When ViT generates patches, the patches are non-overlapping, and thus, at a time pixels in one patch are only able to attend to each other and can’t attend to other patches.
    \item Since images have a large number of features, the distribution of attention scores for these features often becomes too smooth and the goal of attention to focus on certain tokens is lost.
\end{itemize}
We present some techniques to deal with these issues in the following sections.

\subsection{Shifted Patch Tokenization (SPT)}
SPT aims to alleviate the problem of non-overlapping patches. The idea is to consider the interactions between neighbouring pixels during token creation. The receptive fields of tokens are determined by tokenization. The patch generation process in ViT is similar to the convolution operation where the kernel size and stride is equal to the patch size of ViT. But the receptive field of a standard ResNet50 model is about 30 times larger than ViT, hence there is lack of locality inductive bias. 

To resolve this, SPT tries to increase the receptive field. First, the input image is shifted by half the patch size in 4 directions. Then these 4 images are cropped to the same size as the original image and the remaining pixels are padded with zeros. Then all the cropped images are concatenated with the original image. These concatenated features are finally divided into non-overlapping patches, which are flattened for input to the model. The flattened patches are converted into tokens through layer normalization and projection.

\subsection{Rotary Position Embeddings (RoPE)}
As mentioned earlier, the ViT input embeddings are a combination of patch embeddings and positional embeddings. The interactions between tokens can be enhanced by changing the way patch embeddings are generated using SPT. RoPE can then enhance the positional embeddings. The standard ViT uses absolute embeddings where each patch, along with its position is encoded into a single embedding. RoPE suggests using relative positions instead \cite{jeevan2021vision}, such that we can have information about what part is more important. The difference between absolute positions of two tokens is encoded into a single embedding. This ensures that as the relative distance between two patches increases, the RoPE distance decreases, which is desired in case of images.

\subsection{Locality Self-Attention (LSA)}
Another technique that we use is an enhancement to the attention mechanism. The equations below summarize the attention mechanism of the standard ViT.
\begin{equation}\label{eqn:one}R(x) = xE_q(xE_k)^T\end{equation}
\begin{equation}\label{eqn:two}SA(x) = softmax(R/\sqrt{d_k})xE_v\end{equation}
The similarity matrix R in equation \ref{eqn:one} is obtained after matrix multiplication of Q and K vectors, multiplied by their weight matrices. The diagonal tokens in R represent the self-token relations and the other entries represent the inter-token relations. The final attention scores are obtained by taking a softmax of the similarity matrix vector divided by a temperature term, which is the square root of the dimensions of the key. However, the dot product of self-tokens of query and key becomes too large and dominates the other terms. Hence, equation \ref{eqn:two} would give relatively high scores to self-token relations and small scores to inter-token relations. The second problem is that the $\sqrt{d_k}$ term is added to avoid softmax from generating very small gradients. But in case of images, ${d_k}$ can be very large and cause smoothing of attention scores.

So, to avoid this, two techniques are proposed \cite{lee2021vision}. The first is diagonal masking. Since self-tokens smooth out attention, it is best to remove those so that scores are more uniformly distributed over the inter-token relations. The other solution is to use a temperature parameter that is learnt during training. This new temperature is found to be lower, and hence, doesn’t smooth out attention scores.

\subsection{Adversarial Loss}\label{disc}
\begin{figure}[h]
\centering
\includegraphics[scale=0.4]{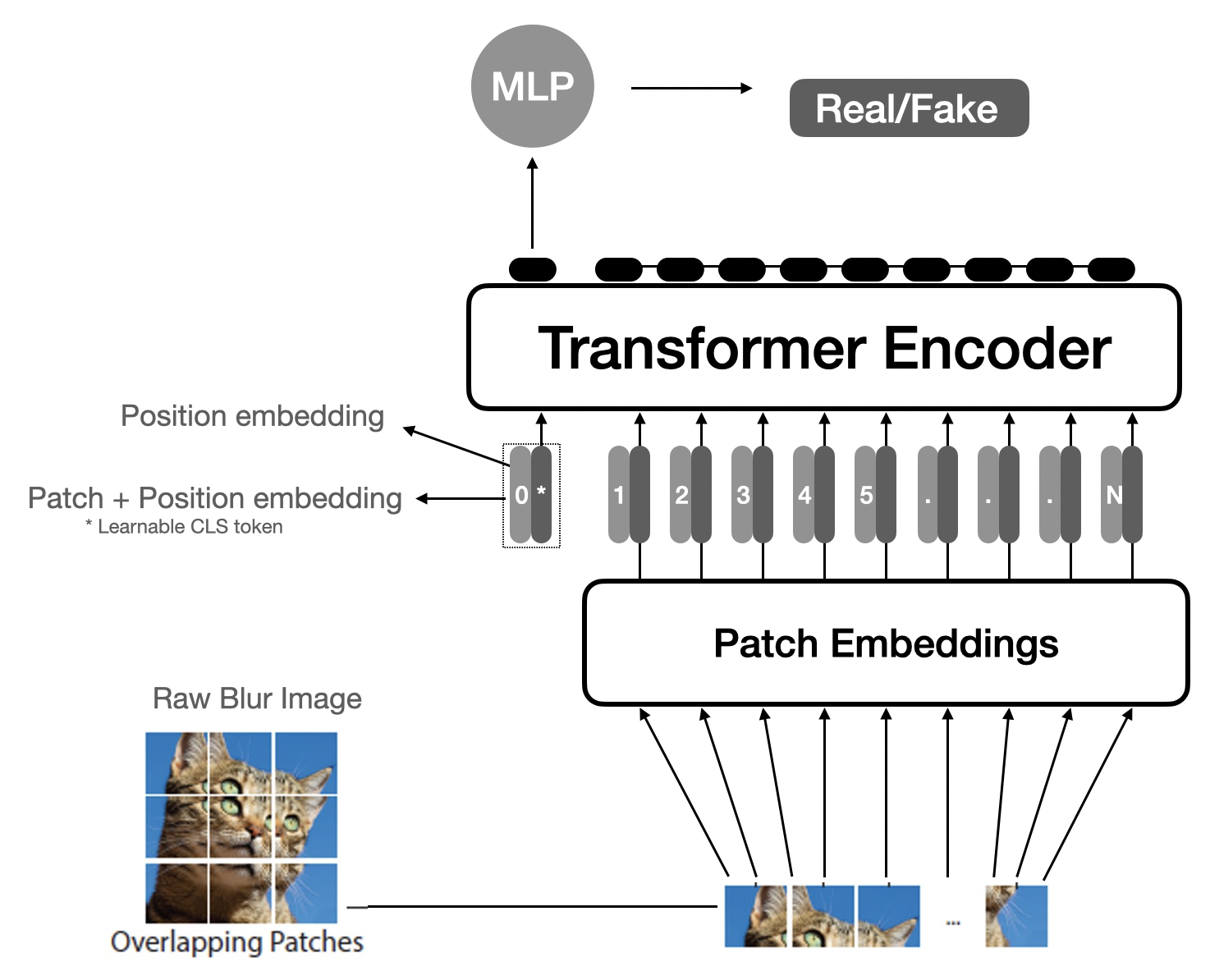}
\caption{Architecture of the discriminator}
\label{fig:disc}
\end{figure}

The model is trained using a combined SSIM loss and adversarial cross entropy loss \cite{goodfellow2014generative}. Since ViT-based generator is unstable when training with CNN-based discriminator, the proposed model uses a ViT-based discriminator \cite{lee2021vitgan} which has a structure that assimilates the ViT-generator. Additional regularization techniques, such as overlapping input image patches \cite{lee2021vitgan} and L2 attention \cite{durall2021combining} are used to further stabilize the training process. The architecture of the discriminator is shown in Fig \ref{fig:disc}.

\section{Experimental Setup}

This section describes the details of the proposed architecture such as dataset used, metrics measured and hardware/ memory constraints.
\subsection{Dataset}
The proposed ViT architecture and the baseline (U-Net) architecture were evaluated based on the Tiny ImageNet dataset, which is a subset of the ImageNet dataset from the famous ILSVRC challenge \cite{ILSVRC15}. The dataset contains 100,000 images of 200 classes downsized to 64×64 images. Due to time and computational resource constraints, subsets of Tiny ImageNet training and testing datasets were used. 20,000 images were used for training and 4,000 images were used for testing. The images were also converted into grayscale for training and evaluation. Data augmentation techniques, such as mirroring and randomly rotating some images by multiples of 90$^{\circ}$, were used to ensure that the model performs well regardless of image orientation. 

\subsection{Evaluation System and Metrics}
All the models were evaluated based on the two image reconstruction tasks using three metrics - Peak Signal-to-Noise Ratio (PSNR), Structural Similarity Index (SSIM) and Normalized Mean Square Error (NMSE). For the image denoising task, the images were blurred by adding Gaussian noise with a variance of 0.05 and the systems were evaluated based on the ability to reconstruct a denoised image. For the image inpainting task, rows of pixels in the images were randomly blacked out and the models were evaluated based on the ability to reconstruct the images without the black-out pixels. 

Regarding the evaluation metrics, PSNR was used to compare the maximum power of a signal and the power of the noise in the images. In particular, PSNR is sensitive to Gaussian additive noise \cite{hore2010image}. SSIM and NMSE both measure the similarity between the original image and the reconstructed image. SSIM quantifies the visibility of differences between a distorted image and a reference image based on properties of the human visual system \cite{rehman2011ssim}. A higher SSIM or lower NMSE is preferred and indicates that the reconstructed image is closer to the original image before adding noise or masks. 

\subsection{Compute hardware and memory used}
All the models were run on a 8 core NVIDIA RTX A6000 GPU with 32G memory and different batch sizes were chosen for different experiments such that the maximum number of images could fit into each batch, given the memory. Based on our testing, the models are not sensitive to batch sizes and changing batch size does not cause noticeable difference in metrics.

\section{Experiments and Results}\label{sec:results}
Four experiments were performed to examine the performance of the benchmark U-Net architecture, ViT architecture and individual enhancement techniques, and to explore the optimal combination of enhancements proposed. The experiments are discussed in the following sections. 

\subsection{Experiment 1: Comparing U-Net and ViT}
In the first experiment, we compared the performance of vanilla ViT model and the benchmark U-Net model to evaluate whether the image reconstruction tasks with ViT architecture can match or outperform the conventional U-Net model. Based on the results shown in Table \ref{tab:individualComparison}, vanilla ViT architecture outperforms the U-Net architecture on the three metrics for both denoising and inpainting task by huge margins. Thus, the vanilla ViT for image reconstruction forms a good baseline for further experiments.

\begin{table*}[!htbp]
\centering
\caption{Comparison among baseline U-Net model, Vanilla ViT and Vanilla ViT with individual enhancement techniques for denoising and inpainting tasks}
\label{tab:individualComparison}
\def\arraystretch{1.3}
\begin{tabular}{c|c|c|c|c|c}
\hline
 {\bfseries Denoising} & {\bfseries U-Net} & {\bfseries Vanilla ViT} & {\bfseries LSA} & {\bfseries SPT} & {\bfseries RoPE}  \\\hline
 PSNR & 21.90 & 27.10 & 27.03 & 27.23 & {\bfseries 27.63}  \\
 SSIM & 75.53\% & 78.95\% & 78.91\% & 79.96\% & {\bfseries 80.86\%}  \\
 NMSE & 3.66\% & 1.04\% & 1.04\% & 0.99\% & {\bfseries 0.90\%} \\\hline
 {\bfseries Inpainting} & {\bfseries U-Net} & {\bfseries Vanilla ViT} & {\bfseries LSA} & {\bfseries SPT} & {\bfseries RoPE}  \\\hline
 PSNR & 21.90 & 24.69 & 24.75 & 24.79 & {\bfseries 25.18} \\
 SSIM & 75.53\% & 79.18\% & 79.38\% & 79.81\% & {\bfseries 80.81\%} \\
 NMSE & 3.66\% & 2.30\% & 2.21\% & 2.21\% & {\bfseries 2.04\%} \\\hline
\end{tabular}
\end{table*}

\subsection{Experiment 2: Using LSA, SPT and RoPE in ViT}
The second experiment was performed to analyze the impact of individual enhancement techniques. As shown in Table \ref{tab:individualComparison}, for LSA, there is no significant difference in metrics before and after adding LSA to the vanilla ViT. It performed slightly better than the vanilla ViT in the inpainting task but underperformed in the denoising task. This might be because the smoothing of attention scores may not be a problem for this dataset since the images are scaled down to 64x64 pixels. LSA might be able to show more improvement for another dataset with larger images. For SPT and RoPE, the metrics for both enhancement methods outperformed the vanilla ViT in denoising and inpainting tasks. RoPE model showed the best results out of all the enhancement techniques. This means that the ViT benefited the most from changing the positional embedding.

\begin{table*}[!htbp]
\centering
\caption{Comparison among Vanilla ViT and Vanilla ViT with individual attention enhancement techniques after incorporating the discriminator based on denoising and inpainting tasks}
\label{tab:withDiscriminator}
\def\arraystretch{1.3}
\begin{tabular}{c|c|c|c|c}
\hline
 {\bfseries Denoising} & {\bfseries Vanilla ViT} & {\bfseries LSA} & {\bfseries SPT} & {\bfseries RoPE}  \\\hline
 PSNR & 27.10 (-0.00) & 27.03 (+0.00) & 27.36 (+0.12) & {\bfseries 27.69 (+0.06)}  \\
 SSIM & 79.06\% (+0.10\%) & 79.08\% (+0.17\%) & 80.26\% (+0.30\%) & {\bfseries 80.99\% (+0.13\%)}  \\
 NMSE & 1.03\% (-0.01\%) & 1.04\% (+0.00) & 0.96\% (-0.03\%) & {\bfseries 0.89\% (-0.01\%)} \\\hline
 {\bfseries Inpainting} & {\bfseries Vanilla ViT} & {\bfseries LSA} & {\bfseries SPT} & {\bfseries RoPE}  \\\hline
 PSNR & 24.81 (+0.12) & 24.79 (+0.04) & 24.96 (+0.16) & {\bfseries 25.19 (+0.95)} \\
 SSIM & 79.36\% (+0.18\%) & 79.50\% (+0.12\%) & 80.03\% (+0.22\%) & {\bfseries 80.86\% (+0.05\%)} \\
 NMSE & 2.20\% (-0.10\%) & 2.23\% (+0.03\%) & 2.13\% (-0.08\%) & {\bfseries 2.05\% (+0.01\%)} \\\hline
\end{tabular}
\end{table*}

\subsection{Experiment 3: Adding adverserial loss function}
After experimenting with individual enhancements, we conducted experiments on the discriminator. The setup of the third experiment was similar to the second experiment, except that the discriminator described in section \ref{disc} was added to the architecture and the network was trained based on adversarial loss. As shown in Table \ref{tab:withDiscriminator}, the discriminator improved the performance of the models overall, but the improvement is marginal. This means that this dataset does not benefit much from the addition of the discriminator. Also, similar to the last experiment, the model with RoPE and discriminator outperformed other models in all cases.

\subsection{Experiment 4: Combination of techniques}
\begin{figure}[h]
\centering
\includegraphics[scale=1.1]{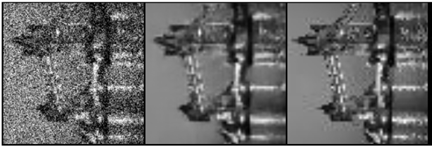}
\caption{Denoised image reconstructed using the proposed framework. Image with noise (left), reconstructed image (middle), original image (right)}
\label{fig:denoiseRecons}
\end{figure}

\begin{figure}[h]
\centering
\includegraphics[scale=1.1]{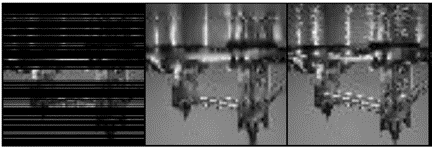}
\caption{Inpainted image reconstructed using the proposed framework. Image with paint (left), reconstructed image (middle), original image (right)}
\label{fig:inpaintRecons}
\end{figure}

The last experiment was conducted to explore the optimal combinations of enhancement techniques so as to propose one final model for each task. For image denoising, the best results were seen when LSA, SPT, RoPE and discriminator were used together. However, there was only marginal improvement in comparison to using only RoPE. For image inpainting, the combination of all enhancement techniques also resulted in the best metrics. Note that other combinations were also tested but only the best performing ones have been listed in Table \ref{tab:combination}. The images generated from the best combination for both the tasks are shown in Fig \ref{fig:denoiseRecons} and \ref{fig:inpaintRecons}.

\begin{table*}[!htb]
\centering
\renewcommand{\arraystretch}{1.3}
\centering
\caption{Comparison among combinations of various enhancement techniques}
\label{tab:combination}
\begin{tabular}{p{0.12\textwidth} | p{0.13\textwidth} | p{0.14\textwidth} | p{0.14\textwidth} | p{0.14\textwidth} | p{0.14\textwidth}}
\hline
{\bfseries Denoising} & {\bfseries ROPE} & {\bfseries ROPE, SPT} & {\bfseries ROPE, LSA, SPT} & {\bfseries LSA, SPT, Discriminator} & {\bfseries LSA, SPT, ROPE, Discriminator}  \\\hline
PSNR & {\bfseries 25.18} & 25.12 & 25.13 & 24.89 & {\bfseries 25.18} \\
SSIM & 80.81\% & 80.88\% & 80.95\% & 79.93\% & {\bfseries 80.96\%} \\
NMSE & {\bfseries 2.04\%} & 2.05\% & 2.05\% & 2.16\% & 2.05\% \\\hline
{\bfseries Inpainting} & {\bfseries ROPE} & {\bfseries ROPE, SPT} & {\bfseries ROPE, LSA, SPT} & {\bfseries LSA, SPT, Discriminator} & {\bfseries LSA, SPT, ROPE, Discriminator}  \\\hline
PSNR & 24.75 & 24.65 & 24.65 & 24.43 & {\bfseries 25.18} \\
SSIM & 79.35\% & 79.36\% & 79.33\% & 78.40\% & {\bfseries 80.96\%} \\
NMSE & 2.22\% & 2.29\% & 2.27\% & 2.39\% & {\bfseries 2.05\%} \\\hline
\end{tabular}
\end{table*}

\section{Limitations and Future Scope}\label{sec:limitations}
\begin{figure}[h]
\centering
\includegraphics[scale=0.5]{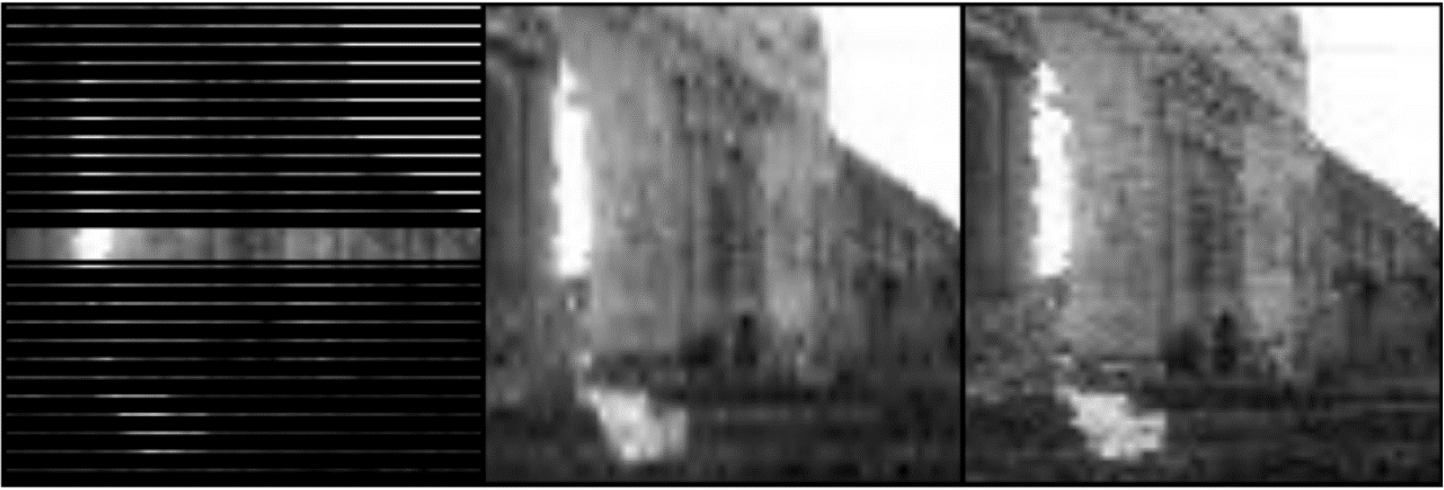}
\caption{Illustration of image after inpainting (left), reconstructed image (middle) and original image (right)}
\label{fig:limitations}
\end{figure}

Based on a thorough analysis of the proposed techniques, we identified some limitations of our project that could be enhanced in the future. On comparing the original and reconstructed image, we noticed that the proposed system tends to overly smoothen out some of the image details, such as the smudges the edges. See figure \ref{fig:limitations}  for illustration. This suggests that the current system is not perfect in distinguishing between image details and noise in the image and further enhancements may be required.

Furthermore, to ensure that the experimental analysis can be generalized well to other cases, more comprehensive experiments need to be performed. For example, it would be beneficial to perform experiments on more datasets and colored images. Datasets with larger image sizes may help uncover some more interesting details, such as LSA can be expected to perform better for larger images. Also, the performance of the models can be examined based on a wider variety of image reconstruction tasks such as adding different noise filter, image super-resolution, deblurring etc.

It would also be interesting to explore in depth how the different techniques interact with each other as part of the whole architecture and whether two techniques counteract the effect of each other.

\section{Conclusion}\label{sec:conclusion}
In this project, we demonstrated that ViT-based architecture can be used for image reconstruction tasks such as image denoising and inpainting, and potentially outperforms the conventional U-Net architecture. We proposed a novel architecture involving four enhancements to various components of the ViT - LSA, SPT, RoPE and adding a discriminator. We showed that the latter three techniques improve the image reconstruction capability of the ViT architecture on the Tiny Imagenet dataset. Based on the experiments, out of all the techniques, RoPE performed best individually and the combination of all enhancement techniques resulted in the best performance for both denoising and inpainting tasks. The proposed architecture can be used for a wide variety of applications where reconstructing noiseless images is critical.

\section{Acknowledgement}\label{sec:ack}
The authors thank the CSC2547 course staff, including Prof. Anthony Bonner for his constant guidance and motivation for the project. The authors are also grateful to the Department of Computer Science, University of Toronto for providing computational resources, without which the project would not have been possible.


\bibliography{irvit}

\bibliographystyle{plainnat}



\end{document}